%% file: ijcai24.tex
\documentclass{article}
\usepackage{ijcai24}
\usepackage{graphicx}
\usepackage{times} 
\usepackage{soul}
\usepackage{url}
\usepackage[hidelinks]{hyperref}
\usepackage[utf8]{inputenc}
\usepackage[small]{caption}
\usepackage{amsmath}
\usepackage{amsthm}
\usepackage{booktabs}
\usepackage{algorithm}
\usepackage[switch]{lineno}
\usepackage{xcolor}
\usepackage{array}
\usepackage{siunitx}
\usepackage{amssymb}
\usepackage{amsfonts}
\usepackage{algpseudocode}
\usepackage{adjustbox}
\usepackage{float}
\usepackage{tabularx}
\usepackage{arydshln}
\usepackage{bm}

\title{BAISeg: Boundary Assisted Weakly Supervised Instance Segmentation}

\author{
    Tengbo Wang, Yu Bai
    \affiliations
    HEBUSTS
    \emails
     tbwang.research@gmail.com , baiyu@hebust.edu.cn
}

\begin{document}

\maketitle

\begin{abstract}
How to extract instance-level masks without instance-level supervision is the main challenge of weakly supervised instance segmentation (WSIS). Popular WSIS methods estimate a displacement field (DF) via learning inter-pixel relations and perform clustering to identify instances. However, the resulting instance centroids are inherently unstable and vary significantly across different clustering algorithms. In this paper, we propose Boundary-Assisted Instance Segmentation (BAISeg), which is a novel paradigm for WSIS that realizes instance segmentation with pixel-level annotations. BAISeg comprises an instance-aware boundary detection (IABD) branch and a semantic segmentation branch. The IABD branch identifies instances by predicting class-agnostic instance boundaries rather than instance centroids, therefore, it is different from previous DF-based approaches. In particular, we proposed the Cascade Fusion Module (CFM) and the Deep Mutual Attention (DMA) in the IABD branch to obtain rich contextual information and capture instance boundaries with weak responses. During the training phase, we employed Pixel-to-Pixel Contrast to enhance the discriminative capacity of the IABD branch. This further strengthens the continuity and closedness of the instance boundaries. Extensive experiments on PASCAL VOC 2012 and MS COCO demonstrate the effectiveness of our approach, and we achieve considerable performance with only pixel-level annotations. The code will be available at https://github.com/wsis-seg/BAISeg.

\end{abstract}

\section{Introduction}

\begin{figure}[t]
    \centering
    \includegraphics[width=\linewidth]{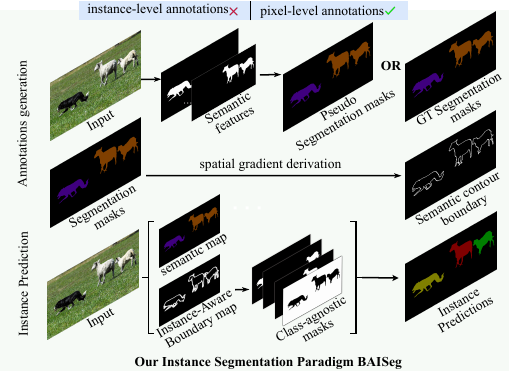}
    \caption{The diagrams illustrate our proposed approach BAISeg, with both label generation and the prediction of instance segmentation masks. Compared to the existing methods, ours does not require instance-level annotations and can be derived from existing semantic segmentation masks, resulting in higher efficiency. Samples are from the PASCAL VOC 2012 dataset~\protect\cite{everingham2010pascal}.}
    \label{fig:paradim}
\end{figure}

Instance segmentation is a task that jointly estimates pixel-level semantic classes and instance-level object masks, and it has made significant progress with the assistance of large datasets. However, it is highly time-consuming and costly to obtain pixel-level and instance-level annotations. To solve the expensive labeling problem, weak annotations such as box-level, point-level, and image-level annotations have been utilized in instance segmentation. These weakly-supervised instance segmentation (WSIS) methods can be roughly categorized into two approaches: the top-bottom approach and the bottom-up approach.

The top-down approach first localizes the region of instances and then extracts instance masks. For instance, \cite{zecheng2023CIM,ge2022empaste} use image-level annotations and derive instance regions from proposals, therefore simplifying WSIS as a classification problem. \cite{lan2021discobox,tian2021boxinst,li2022box2mask} directly utilize box-level annotations and guide the network to learn instance segmentation by a delicate architecture and loss design. Instance-level annotations are used either indirectly or directly in these methods.
In contrast, bottom-up methods require learning relationships between pixels from pre-computed or known instance cues.
For instance, \cite{ahn2019weakly,kim2022beyond} use pseudo segmentation labels that are derived from Class-Activation Maps and Weakly-Supervised Semantic Segmentations to pinpoint instance locations, whereas \cite{liao2023attentionshift,kim2022beyond,cheng2022pointly} utilize point supervision of instances directly.

These methods generate displacement fields (DF) by learning the spatial relationship between pixels and further derive instance masks through post-processing algorithms such as clustering or pixel grouping.
A common problem of the above works is that the centroids identified via clustering inherently lack stability (even ground truth centroids are provided) and are especially vulnerable to variations in clustering algorithms. 

To address the above limitations, we proposed Boundary-Assisted Instance Segmentation (BAISeg), which is a new paradigm that carries out instance segmentation by using pixel-level annotations only.  As shown in the ``Instance Prediction'' of Figure \ref{fig:paradim}, BAISeg comprises an instance-aware boundary detection (IABD) branch and a semantic segmentation branch. The IABD branch follows a top-down approach to extract instance masks by predicting class-agnostic instance boundaries. The semantic segmentation branch is responsible for deriving semantic masks. Finally, the semantic masks and the class-agnostic instance masks are combined to obtain the instance segmentation results. 

In particular, we proposed the Cascade Fusion Module (CFM) and the Deep Mutual Attention(DMA) in the IABD branch to obtain rich contextual information and capture instance boundaries with weak responses. During the training phase, we employed Pixel-to-Pixel Contrast to tackle the semantic drift problem~\cite{kim2022beyond} and strengthened the continuity and closedness of the instance boundaries.

As depicted in ``annotations generation'' of Figure~\ref{fig:paradim}, there are two sources for the semantic segmentation masks. This makes our approach very flexible. Even without off-the-shelf proposal techniques, our approach achieves a competitive performance of 62.0\% {$mAP_{50}$} on VOC 2012 and 33.6\% {$mAP_{50}$} on the COCO Test-Dev. The main contributions of this paper are as follows:

\begin{itemize}
    \item  We propose a novel WSIS paradigm -- BAISeg. BAISeg utilizes a top-down approach to instance mask extraction, which does not require any instance-level annotations and avoids the limitations of relying on proposal algorithms or {estimated centroids}.
    \item To extract instance masks, we propose the CFM module. CFM decouples the boundary detection branch and segmentation branch. In CFM, we further designed DMA to capture instance boundaries with weak responses.
    \item To tackle the semantic drift problem, we introduced Pixel-to-Pixel Contrast to WSIS with a weighted contrast loss function and further improves the perception and closedness of instance boundaries.
\end{itemize}

\section{Related Work}


\subsection{Instance Edge Detection}

Instance segmentation has made significant progress with the help of large datasets. However, the segmentation result of instance boundaries is not satisfactory. Recent works try to utilize the edge information for improvement. SharpContour \cite{zhu2022sharpcontour} proposes a contour-based boundary refinement method. STEAL \cite{acuna2019devil} designed a unified quality metric for mask and boundary to better integrate edge detection and instance segmentation tasks. InstanceCut \cite{kirillov2016instancecut} proposed a novel MultiCut formulation, deriving the optimal partitioning from images to instances through semantic segmentation and instance boundaries. Our work is most similar to that of InstanceCut, but different from all the aforementioned works.First, we achieve segmentation with only pixel-level annotations. Second, our method does not rely on the setting of MultiCut~\cite{Chopra1993ThePP} and uses a very simple and effective method for instance mask extraction.

\subsection{Weakly-Supervised Instance Segmentation}

The primary challenge in Weakly Supervised Instance Segmentation (WSIS) is to obtain instance-level information from weakly annotated labels. To solve this problem, PRM \cite{zhang2019reliability} introduced Peak Response Maps from existing proposals algorithm \cite{Pont_Tuset_2017} to select appropriate segment proposals. By learning the pixel-to-pixel relationships, IRNet \cite{ahn2019weakly} estimates the image's displacement fields to extract instance masks. BESTIE \cite{kim2022beyond} determines the centroid of instances by combining the centroid network and displacement field and then extracts instance masks using clustering algorithms. 
Point2Mask \cite{li2023point2mask} uses the instance GT points for supervision and utilizes OT theory to extract pseudo-labels for panoptic segmentation from semantic segmentation and instance boundaries. CIM \cite{zecheng2023CIM} alleviates the adverse effects resulting from redundant segmentation, building upon existing proposals. A detailed comparison of the above methods reveals the following issues: (1) Methods based on point supervision indirectly or directly use instance-level annotations. (2) Methods based on ready-made proposals are heavily dependent on existing proposal algorithms. (3) Estimated centroids are particularly susceptible to changes in clustering algorithms. 

In this paper, we propose BAISeg to extract class-agnostic instance masks from instance boundary maps. BAISeg uses only pixel-level supervision and does not rely on any proposal algorithms or centroid estimation.

\subsection{Contrastive Learning}

Contrastive learning is an unsupervised learning method that aims to learn data representations by maximizing the similarity between related samples and minimizing the similarity between unrelated samples. A popular contrastive learning loss function is known as InfoNCE \cite{he2020momentum}. Recent works \cite{wang2021exploring,zhao2021contrastive} applied contrastive learning to dense pixel prediction tasks to increase intra-class compactness and inter-class separability. Inspired by these works, we introduce Pixel-to-Pixel contrastive learning in the training of BAISeg, effectively enhancing the discriminative ability of the IABD branch.

\section{Proposed Method}

\begin{figure*}[t]
    \centering
    \includegraphics[width=\linewidth]{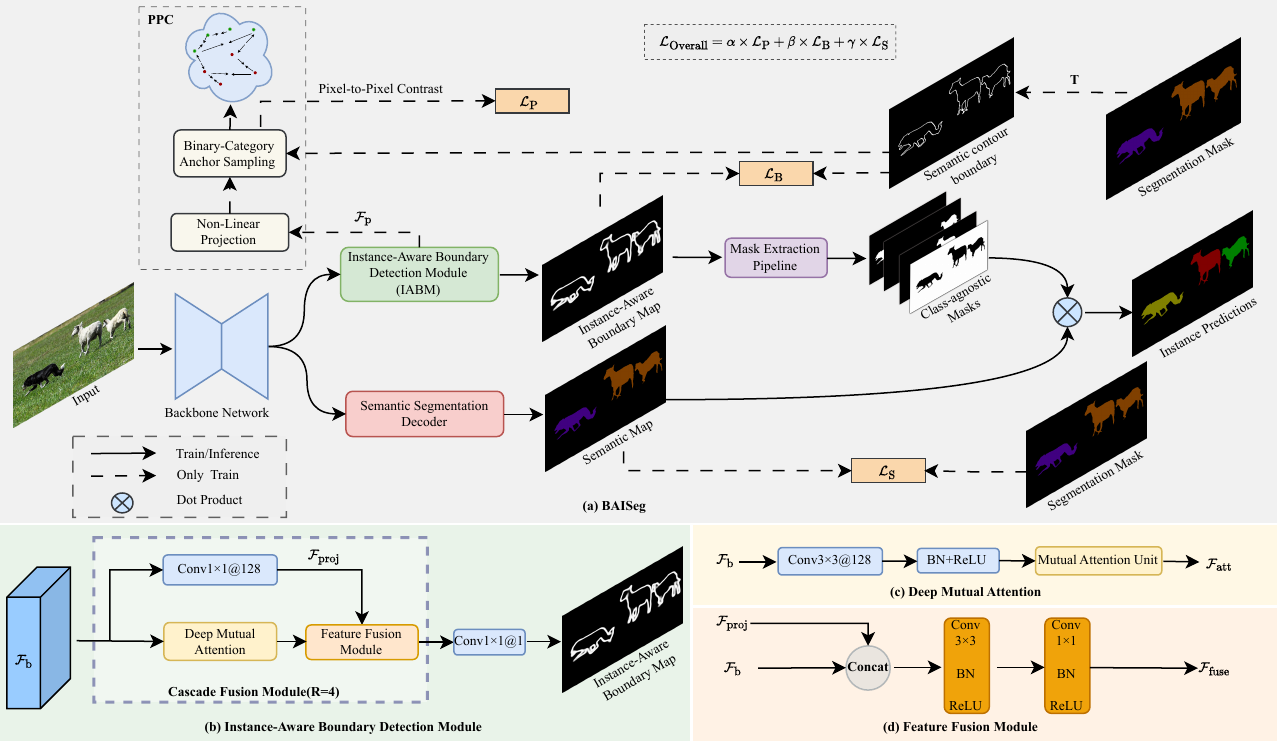}
    \caption{Our proposed BAISeg architecture (a) mainly consists of two parallel branches with a shared backbone: the IABD (b) branch and the semantic segmentation branch. The IABD branch determines the boundaries of instances by predicting instance-aware boundary maps and extracts class-agnostic masks via the Mask Extraction Pipeline. The semantic segmentation branch predicts the semantic maps of instances. Instance segmentations are derived by combining the semantic maps and class-agnostic instance masks. The entire network is optimized by minimizing the $\mathcal{L}_\text{P}$, $\mathcal{L}_\text{B}$, and $\mathcal{L}_\text{S}$ losses on pixel-level annotations. The $T$ function is used to extract semantic contour edge labels from the segmentation masks by spatial gradient deriving. This sample is taken from the validation set of PASCAL VOC 2012.}
    \label{fig:overview}
\end{figure*}

\subsection{Preliminary and Overview}
Given a training image ${X} \in \mathbb{R}^{H \times W \times 3}$ with both pixel-level annotation $\hat{{X}}_{{P}} \in \mathbb{R}^{H \times W \times 1}$ and instance-level annotation $\hat{{X}}_{{I}} \in \mathbb{R}^{H \times W \times 1}$,  the objective of instance segmentation model is to predict an instance-level object mask ${M}_{\text {ins }} \in \mathbb{R}^{H \times W \times 1}$ with pixel-level semantic classes. Our BAISeg also shares the same goal.
As shown in Figure \ref{fig:overview}(a), BAISeg is a one-stage framework that mainly has four components:{Semantic Segmentation Decoder},{Instance-Aware Boundary Detection Module(IABM)}, {Pixel-to-Pixel Contrast(PPC)}, {and a Mask Extraction Pipeline (MEP)}. 
The model has two parallel branches. The IABD branch is responsible for distinguishing different instances belonging to the same class by predicting an Instance-Aware Boundary map. Another branch of semantic segmentation aims to obtain the semantic map of the image. Finally, instance segmentation is achieved by combining semantic maps and instance masks.

\subsection{Instance Boundary Detection Branch}

We propose a novel instance mask extraction method that accomplishes instance segmentation prediction using only pixel-level annotations. In particular, our method identifies instances by instance-aware boundary prediction rather than centroid estimation, therefore does not suffer from instable centroid predictions.

\subsubsection{Branch Structure}
\label{sec:network}

To obtain accurate instance cues and differentiate different instances of the same class, we propose the IABM as shown in Figure \ref{fig:overview} (b) (introduced in Sec. \ref{sec:IABM}), PPC (introduced in Sec. \ref{sec:PPC}), and MEP (introduced in Sec. \ref{sec:MEM}).

An input \(X\) is fed to a backbone network (e.g., HRNet-W48 \cite{sun2019high}) to extract the feature map \(\mathcal{F}_{b} \in \mathbb{R}^{H \times W \times C}\). Then \(\mathcal{F}_{b}\) is forwarded to the IABM to obtain the instance-aware boundary map \(B\) and the boundary feature map \(\mathcal{F}_{p}\). \(B\) is the output of the IABM and is processed by the MEP to extract class-agnostic instance masks. \(\mathcal{F}_{p}\) is extracted from the \(R\) cascaded CFM blocks and is used in pixel-to-pixel contrast to improve the discriminative power of the IABD branch. This further strengthens the continuity and closedness of the instance boundaries.


\subsubsection{Loss Fuction}

We employ the loss function proposed in \cite{xie2015HED} for training boundary maps. Given a boundary map \(B\) and its corresponding ground truth \(Y\), the loss $\mathcal{L}_\text{B}$ is calculated as
\begin{equation}
\begin{aligned}
\mathcal{L}_\text{B}(B, Y) &= -\sum_{i, j} \left(Y_{i, j} \alpha \log \left(B_{i, j}\right)\right.\\
& \quad + \left.(1 - Y_{i, j})(1 - \alpha) \log \left(1 - B_{i, j}\right)\right),
\end{aligned}
\end{equation}
where \(B_{i, j}\) and \(Y_{i, j}\) are the \((i, j)^{th}\) elements of matrices \(B\) and \(Y\), respectively.
Moreover, \(\alpha = \frac{\left|Y^{-}\right|}{\left|Y^{-}\right| + \left|Y^{+}\right|}\), where \(\left|\cdot\right|\) denotes the number of pixels.

\subsection{Instance-Aware Boundary Detection Module}
\label{sec:IABM}

The IABD branch aims to capture instance-aware boundary maps using backbone features, where the clarity and closedness of the boundaries largely limit the performance of instance mask extraction. We designed Instance-Aware Boundary Detection (IABM) to tackle this issue. As shown in Figure \ref{fig:overview}(b), The IABM consists of R Cascade Fusion Modules (CFM) and a standard $1 \times 1$ convolutional layer. 

The design of the IABM is based on two observations. Firstly, the class-agnostic boundary detection branch and the semantic segmentation branch are tightly coupled, and their mutual interaction limits the performance of WSIS. Comprising a cross-layer connection and feature mapping operations, the CFM enables the IABD branch to learn independent and multi-scaled edge feature representations. Secondly, using semantic contour labels alone is insufficient for estimating the clear boundaries of each instance, for which we designed the DMA to focus on specific task features and capture instance boundary information with weak responses. 

\subsubsection{Cascade Fusion Module}

We designed CFM based on two observations: By a cascaded design, the network can learn task-specific local representations in each stage. Secondly, a cascaded network extracts and refines features gradually, which helps capture multi-scaled edge information. As shown in Figure \ref{fig:overview}(b), CFM comprises three components: DMA, $1 \times 1$ convolution, and FFM. Notably, DMA can be substituted by any other convolutional layer.

\subsubsection{Deep Mutual Attention}
\label{subsec:dam}

\begin{figure}[t]
    \centering
    \includegraphics[width=\linewidth]{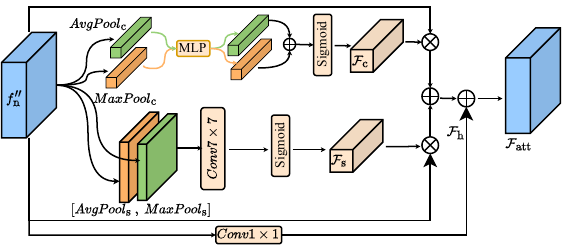}
    \caption{Mutual Attention Unit.}
    \label{fig:DA}
\end{figure}

As a submodule of CFM, {Deep Mutual Attention (DMA)} aims to capture weakly responsive instance boundary information from deeply fused features.
As shown in Figure \ref{fig:overview}(c), DMA consists of two parts: the embedding operation and the {Mutual Attention Unit (MAU)}. The embedding operation comprises $3 \times 3$ convolutional layers followed by Batch Normalization (BN) and ReLU operations. The advantages of DMA encompass two aspects. Firstly, DMA employs the MAU to effectively capture attention features at various stages. Secondly, it integrates these attention features at different stages to capture edge features with weak responses.

The MAU is inspired by the work \cite{tian2021learning}. Notably, we introduced a $1 \times 1$ convolutional layer and learnable attention weights $\beta$ to the MAU, aiming to dynamically adjust the level of attention for the fused features in each CFM while preserving the original feature information. Figure \ref{fig:DA} shows the detailed structure of the MAU. The top and bottom branches are channel-wise and spatial-wise attention blocks, respectively. Given input features $f_n{ }^{\prime \prime}$, we compute the channel-wise attention features $\mathcal{F}_c$ as follows:
\begin{equation}
\mathcal{F}_c\!=\!\sigma\!\left(\operatorname{MLP}\!\left(\operatorname{AvgPool}_c\left(f_n{ }^{\prime \prime}\right)\right)\!\!+\!\!\operatorname{MLP}\!\left(\operatorname{MaxPool}_c\!\left(f_n{ }^{\prime \prime}\right)\right)\right)\!\!,\!\!
\end{equation}
where $\mathrm{MaxPool}_c$ and $\mathrm{AvgPool}$ are the two channel-wise pooling operations, and MLP is the multi-layer perception with one hidden layer to generate the attention features. We also compute the spatial-wise attention features $\mathcal{F}_s$ as:
\begin{equation}
\mathcal{F}_s=\sigma\left(\operatorname{Conv}_{7 \times 7}\left(\left[\operatorname{AvgPool}_s\left(f_n{ }^{\prime \prime}\right) ; \operatorname{MaxPool}_s\left(f_n{ }^{\prime \prime}\right)\right]\right)\right),
\end{equation}
where $\operatorname{Conv}_{7 \times 7}$ is a convolutional layer with kernel size 7 . 
The Features of identity transformation $\mathcal{F}_h$ is computed as:
\begin{equation}
\mathcal{F}_h = \operatorname{Conv}_{1 \times 1}(f_n{ }^{\prime \prime})
\end{equation}
where $\operatorname{Conv}_{1 \times 1}$ is a convolutional layer with kernel size 1 . 
The final attention features $\mathcal{F}_{att}$ are computed as follows:
\begin{equation}
\mathcal{F}_{att}=\mathcal{F}_h+\left(f_n{ }^{\prime \prime} \times \mathcal{F}_c+f_n{ }^{\prime \prime} \times \mathcal{F}_s\right)\times\beta,
\end{equation}
where $\times$ denotes the dot product operation, and $+$ is the element-wise summation operation.

\subsubsection{Feature Fusion Module}
\label{subsec:fuse}

Global features and local features are obtained in the upper and lower branches of CFM respectively. By fusing these features, the Feature Fusion Module (FFM) generates boundary-aware global features and refined local features. The detailed structure of FFM is shown in Figure \ref{fig:overview}(d).

\begin{table*}[t]
	\centering
	\renewcommand\arraystretch{1.2}
	\setlength{\tabcolsep}{9pt}
	\vspace{-1mm}
	\resizebox{\linewidth}{!}{%
		\begin{tabular}{c | c | c | c  c  c  c}
			\hline
			Methods & Backbone & Superv &{$mAP_{25}$} & {$mAP_{50}$}& {$mAP_{70}$} & {$mAP_{75}$}\\
			\hline 
			Mask R-CNN~\cite{he2017mask} & ResNet-101 & $\mathcal{F}$ & \textbf{76.7} & \textbf{67.9} & \textbf{52.5} & \textbf{44.9} \\
			\hline  
                EM-Paste~\cite{ge2022empaste} & ResNet-101 & $\mathcal{I}$  & -- & \textbf{58.4} & \textbf{37.2} & -- \\
			BESTIE~\cite{kim2022beyond} & HRNet-W48 & $\mathcal{I}$ & 53.5 & 41.8 & 28.3 & 24.2\\ 
                CIM~\cite{zecheng2023CIM} & HRNet-W48  & $\mathcal{I}$ & 68.3  & 52.6 & 33.7 & 28.4\\
			BESTIE~\cite{kim2022beyond} & HRNet-W48 & $\mathcal{P}$& 58.6  & 46.7 & 33.1 & 26.3 \\
			AttnShif~\cite{liao2023attentionshift} & ViT-S & $\mathcal{P}$ & \textbf{68.3}  & \textbf{54.4} & -- & \textbf{25.4} \\
			DiscoBox~\cite{lan2021discobox} &  ResNet-101 & $\mathcal{B}$ & 72.8 & 62.2 & 45.5 & 37.5 \\
			SIM~\cite{li2023sim} & ResNet-50  & $\mathcal{B}$ & --  & 65.5 & 35.6 & -- \\
			Box2Mask~\cite{li2022box2mask} & ResNet-50 & $\mathcal{B}$ & \textbf{38.0}  & \textbf{65.9} & \textbf{46.1} & \textbf{38.2}\\
			\cdashline{1-7}[0.8pt/2pt]
			BAISeg (OCRNet) & HRNet-W48 & $\mathcal{P_M}$ & 52.73  & 44.0 & 32.1 & 28.9 \\
			BAISeg (Semantic\_gd) & HRNet-W48 & $\mathcal{P_M}$  & \textbf{55.3}  & \textbf{48.4} & \textbf{36.9} & \textbf{32.3}\\
			\hline
                BESTIE${ }^{\dagger}$~\cite{kim2022beyond} & HRNet-W48 & $\mathcal{P}$ & 66.4 & 56.1 & 36.5 & 30.20\\
                CIM${ }^{\dagger}$~\cite{zecheng2023CIM} & ResNet-50  & $\mathcal{I}$ & 68.7  & 55.9 & 37.1 & 30.9\\
                AttnShif${ }^{\dagger}$~\cite{liao2023attentionshift} & ViT-S & $\mathcal{P}$ & 70.3  & 57.1 & -- & 30.4 \\
                \cdashline{1-7}[0.8pt/2pt]
                BAISeg (Semantic\_gd)${ }^{\star}$ & HRNet-W48 & $\mathcal{P_M}$  & 59.2  & 53.0 & 42.0 & 37.6\\
                BAISeg (Semantic\_gd)${ }^{\star\dagger}$ & HRNet-W48 & $\mathcal{P_M}$  & \textbf{69.4}  & \textbf{62.0} & \textbf{44.0} & \textbf{36.0}\\
                \hline
		\end{tabular}%
            }
 \caption{\fontsize{9}{9}\selectfont Comparison of state-of-the-art methods on the PASCAL VOC 2012validation set. ``--'' represents that the result is not reported in its paper.${ }^{\dagger}$ denotes additional training with Mask R-CNN to refine the prediction.${ }^{\star}$ denotes Inference using the ground truth mask as semantic segmentation. ``Superv.'' represents the training supervision ($\mathcal{F}$: instance-level mask, $\mathcal{I}$: image-level class label, $\mathcal{P}$: point, $\mathcal{B}$: bounding box, $\mathcal{P_M}$: pixel-level mask). BAISeg (OCRNet) and BAISeg (Semantic\_gd) indicate that our BAISeg uses the results of OCRNet \protect\cite{yuan2021segmentation} and ground truth mask as semantic segmentation for training respectively.} 
 \label{tab:sota_pascal}
\end{table*}
%

\subsection{Pixel-to-Pixel Contrast}
\label{sec:PPC}

The limited discriminative power of IABD can lead to issues of instance boundaries not being closed, causing closely connected instances of the same class to be merged into a single instance. To enhance the model's ability to discriminate and perceive instance edges, we introduced the pixel-to-pixel contrast training \cite{wang2021exploring,zhao2021contrastive} in BAISeg. PPC's core idea is to bring similar pixels closer together while repelling dissimilar ones. However, since we use semantic contour labels for sampling, there are a large number of incorrect labels in the background. To mitigate the negative impact of these incorrect labels on the boundary detection branch, we propose a weighted contrastive loss.

\subsubsection{Weighted Pixel-to-Pixel Contrast Loss}

The data samples in our contrastive loss computation are training image pixels.  Given a semantic contour map $Y$, for a pixel $i$ with its boundary pseudo-label $\bar{c}$, the positive samples are those pixels that also belong to the same class $\bar{c}$, while the negatives are the pixels belonging to background class ${c}$. Our supervised, pixel-wise contrastive loss is defined as:
\begin{equation}\small
\!\!\!\!\mathcal{L}_\text{P}\!=\!\frac{1}{|\mathcal{P}_i|}\!\!\sum_{\bm{i}^+\in\mathcal{P}_i\!\!}\!\!\!-_{\!}\alpha\log\frac{\exp(\bm{i}\!\cdot\!\bm{i}^{+\!\!}/\tau)}{\exp(\bm{i}\!\cdot\!\bm{i}^{+\!\!}/\tau)\!
+\!(1-\alpha)\!\!\sum\nolimits_{\bm{i}^{-\!}\in\mathcal{N}_i\!}\!\exp(\bm{i}\!\cdot\!\bm{i}^{-\!\!}/\tau)},\!\!\!
\end{equation}
where $\mathcal{P}_i$ and $\mathcal{N}_i$ denote pixel embedding collections of the positive and negative samples, respectively, for pixel $i$. Moreover, \(\alpha = \frac{\left|Y^{-}\right|}{\left|Y^{-}\right| + \left|Y^{+}\right|}\), where \(\left|\cdot\right|\) denotes the number of pixels. Note that the positive/negative samples and the anchor $i$ are not restricted to originate from the same image.

\subsection{Overall Loss Function} 

The loss function for our semantic segmentation branch is derived from DeepLabv3 \cite{deeplabv3plus2018}, and we denote it as $\mathcal{L}_\text{S}$. The overall loss function of BAISeg can be formulated as follows:
\begin{equation}
\mathcal{L}_\text{Overall}=\alpha\times\mathcal{L}_\text{P}+\beta\times\mathcal{L}_\text{B}+\gamma\times\mathcal{L}_\text{S},
\end{equation}
where $\alpha$, $\beta$, and $\gamma$ represent the weights for the three loss components, and we set $\alpha=0.3$, $\beta=50$, and $\gamma=1$.

\begin{figure*}[t]
    \centering
    \includegraphics[width=\linewidth]{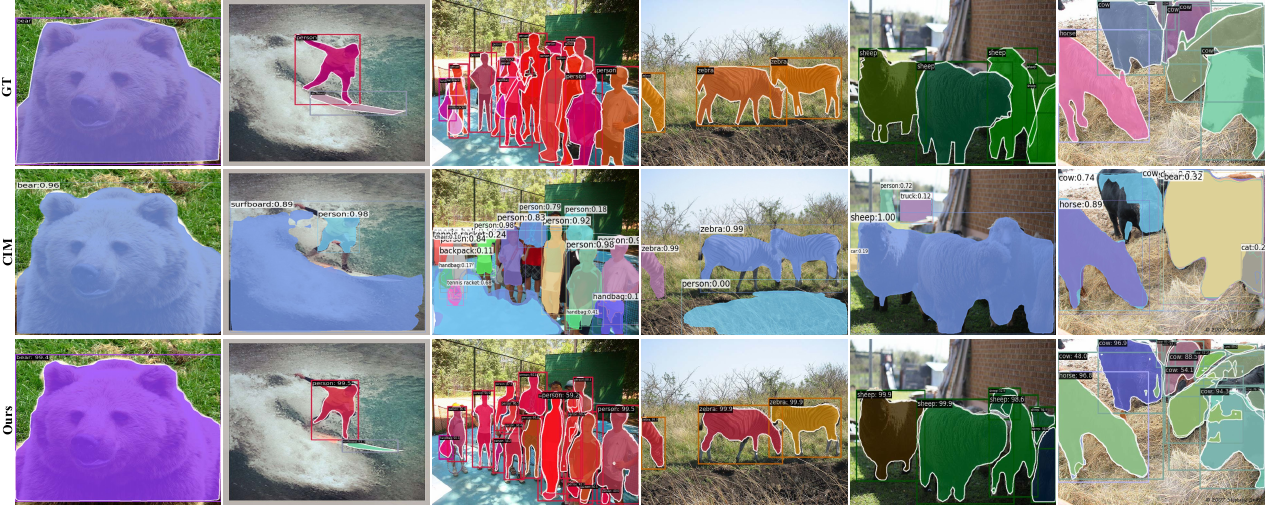}
    \caption{Visualization results on the COCO 2017 dataset. Comparison with CIM \protect\cite{zecheng2023CIM}. }
    \label{fig:contrast_cim}
\end{figure*}

\begin{table}[ht!]
  \centering
  \renewcommand{\arraystretch}{1.2} 
  \begin{adjustbox}{max width=1\linewidth}
    \begin{tabular}{c|c|c|ccc}
      \hline
      Method & Backbone & Superv & $AP$ & {$AP_{50}$} & {$AP_{75}$}  \\
      \hline
      \multicolumn{6}{c}{\textit{\textbf{COCO val2017}}} \\
      \hline
      Mask R-CNN~\cite{he2017mask} & ResNet-101 & $\mathcal{F}$ & 35.4 & 57.3 & 37.5 \\
      CIM~\cite{zecheng2023CIM} & ResNet-50  & $\mathcal{I}$ & 11.9  & 22.8 & 11.1 \\
      CIM${ }^{\dagger}$~\cite{zecheng2023CIM} & ResNet-50  & $\mathcal{I}$ & 17.0  & 29.4 & 17.0 \\
      BESTIE${ }^{\dagger}$~\cite{kim2022beyond} & HRNet-W48 & $\mathcal{P}$ & 17.7 & 34.0 & 16.4 \\
      AttnShif~\cite{liao2023attentionshift} & ViT-S & $\mathcal{P}$ & 19.1 & 38.8 & 17.4 \\
      Box2Mask~\cite{li2022box2mask} & ResNet-50 & $\mathcal{B}$ & 32.2  & 54.4 & 32.8 \\
      \cdashline{1-6}[0.8pt/2pt]
      BAISeg (Semantic\_gd)${ }^{\star}$ & HRNet-W48 & $\mathcal{P_M}$& \textbf{15.8} & \textbf{25.8} & \textbf{15.9} \\
      BAISeg (Semantic\_gd)${ }^{\dagger}$ & HRNet-W48 & $\mathcal{P_M}$ & \textbf{19.1} & \textbf{33.4} & \textbf{19.0} \\
      \hline
      \multicolumn{6}{c}{\textit{\textbf{COCO Test-Dev}}} \\
      \hline
      Mask R-CNN~\cite{he2017mask} & ResNet-101 & $\mathcal{F}$ & 35.7 & 58.0 & 37.8 \\
      CIM${ }^{\dagger}$~\cite{zecheng2023CIM} & ResNet-50  & $\mathcal{I}$ & 17.0  & 29.7 & 17.0 \\
      LIID${ }^{\dagger}$~\cite{liu2020leveraging} & ResNet-101 & $\mathcal{I}$ & 16.0 & 27.1 & 16.5\\
      BESTIE${ }^{\dagger}$~\cite{kim2022beyond} & HRNet-W48 & $\mathcal{P}$& 17.8  & 34.1 & 16.7 \\ 
      AttnShif~\cite{liao2023attentionshift} & ViT-S & $\mathcal{P}$ & 19.1 & 38.9 & 17.1 \\
      \cdashline{1-6}[0.8pt/2pt]
      BAISeg (Semantic\_gd)${ }^{\star\star}$ & HRNet-W48 & $\mathcal{P_M}$& \textbf{13.5} & \textbf{22.8} & \textbf{13.8}   \\
      BAISeg (Semantic\_gd)${ }^{\dagger}$ & HRNet-W48 & $\mathcal{P_M}$ & \textbf{19.0} & \textbf{33.6} & \textbf{19.0}\\
      \hline
    \end{tabular}
  \end{adjustbox}
  \caption{
     \fontsize{9}{9}\selectfont Comparison with the state-of-the-art methods on COCO dataset.${ }^{\dagger}$ denotes additional training with Mask R-CNN to refine the prediction.${ }^{\star}$ denotes Inference using the ground truth mask as semantic segmentation.${ }^{\star\star}$ denotes Inference using the result of mask2former \protect\cite{cheng2022maskedattention} as semantic segmentation . ``Superv.'' represents the training supervision ($\mathcal{F}$: instance-level mask, $\mathcal{I}$: image-level class label, $\mathcal{P}$: point, $\mathcal{B}$: bounding box,$\mathcal{P_M}$: pixel-level mask). 
    BAISeg (Semantic\_gd) means that BAISeg employs the result of the ground truth mask as semantic segmentation.
  }
  \label{tab:sota_coco}
\end{table}

\subsection{Mask Extraction Pipeline}
\label{sec:MEM}
The IABD branch aims to extract class-agnostic instance masks from the instance boundary map. In simple terms, we utilize the well-known watershed algorithm to extract instance masks from the boundary map. However, the instance boundaries generated by the IABD branch have the following issues: (1) roughness of the boundaries, (2) closedness of the boundaries, and (3) holes in the masks. Therefore, we propose a specific Mask Extraction Pipeline (MEP) to deal with these problems.
The whole pipeline can be divided into three stages. In stage one, {Non-Maximum Suppression (NMS) }is applied to process the boundary map to derive thinner boundaries. In stage two, a label map is extracted from the instance boundaries by the Connected-Component Labeling (CCL) algorithm \cite{HE20091977_CCL}.
In the third stage, the input image and label map are combined and then processed by the watershed algorithm \cite{watershed_algorithm} to derive the class-agnostic instance masks. 

To fill the holes in the derived instance masks, we further employed a closing operation. To improve the closedness of the boundaries, the semantic map is utilized as a filter to select only the pixels within the semantic region. We refer to the above additional operations as ``refinement''.

\section{Experiments}
\subsection{Datasets and Evaluation Metrics}

Following previous methods, we demonstrate the effectiveness of the proposed approach on Pascal VOC 2012 \cite{everingham2010pascal} and COCO \cite{lin2014microsoft} datasets. The VOC 2012 dataset includes 10,582 images for training and 1,449 images for validation, comprising 20 object categories. The COCO dataset consists of 115K training, 5K validation, and 20K testing images with 80 object categories.  We evaluate the performance using the mean Average Precision ({$mAP$}) with intersection-over-union (IOU) thresholds of 0.25, 0.5, 0.7, and 0.75 for VOC 2012 and Average Precision ({$AP$}) over IoU thresholds from 0.5 to 0.95 for COCO.

\subsection{ Implementation Details}

We used the PyTorch 1.13 framework with CUDA 11.7, CuDNN 8, and NVIDIA A40 GPUs. We adopt HRNet48 \cite{tian2021learning} as our backbone network. The input image size for training is $416 \times 416$, and we keep the original resolution for evaluation. We train the network with a batch size of 16, the Adam optimizer \cite{kingma2017adam} with a learning rate of $5 \times 10^{-5}$, and polynomial learning rate scheduling \cite{liu2015parsenet}. The total number of training iterations is $6 \times 10^{4}$ for the VOC 2012 dataset and $48 \times 10^{4}$ iterations for the COCO dataset. The training configuration of our Pixel-to-Pixel Contrast follows \cite{wang2021exploring}. Following previous works \cite{rcnn1,zecheng2023CIM}, we also generate pseudo labels from BAISeg for training Mask R-CNN. 

\subsection{Comparison With State-of-the-Art}
\subsubsection{Results on PASCAL VOC 2012}

To illustrate the superior capabilities of BAISeg, we present a comparative analysis with other leading instance segmentation methods on the PASCAL VOC 2012 validation set, as shown in Table \ref{tab:sota_pascal}. Our method is evaluated without any special techniques or tricks. BAISeg experiments on segmentation masks of varying quality, and among these, BAISeg (Semantic\_gd) demonstrated the best performance, achieving an accuracy of 55.32\% {$mAP_{25}$}, 48.40\% {$mAP_{50}$}, and 36.97\% {$mAP_{70}$}, and outperforms methods that rely solely on image-level supervision signals. Compared to approaches based on proposals, point-level, and box-level methods, BAISeg (Semantic\_gd) also achieves competitive results. BAISeg (Semantic\_gd)${}^{\star\dagger}$exceeds methods based on image-level, point-level, and proposals. The performance of {BAISeg} is also competitive compared to fully supervised and box-level methods.

\subsubsection{Results on COCO 2017}
In Table~\ref{tab:sota_coco}, we conduct a comparison of BAISeg with other leading instance segmentation methods on the COCO validation and Test-Dev datasets. To ensure a fair comparison, none of the methods use additional training data. BAISeg (Semantic\_gd)${ }^{\star\dagger}$ has significant improvements over the best image-level methods in terms of {$AP_{50}$}, improving by 4\% and 3.9\% on the validation set and Test-Dev set respectively. Compared to point-level methods, BAISeg performs comparably as well. However, there is still a slight gap when compared to methods trained with stronger localization supervision like Box2Mask and Mask R-CNN. Indeed, our primary focus is on establishing a novel paradigm for WSIS, where quantitative improvement is regarded as a secondary goal.

\subsection{Ablation Study}
We conduct several ablation studies on the Pascal VOC 2012 dataset to evaluate the effectiveness of each component. In these studies, we use HRNet-W48 as the backbone and omit MRCNN refinement to save time.

\subsubsection{Impact of CFM}
\begin{table}[t]
  \centering
  \begin{adjustbox}{max width=1\linewidth}
  \begin{tabular}{cccc|S[table-format=2.1]S[table-format=2.1]}
    \toprule
    CFM & DMA & PPC & Refinement & {$mAP_{50}$} & {$mAP_{75}$}  \\
    \midrule
    $\times$ & $\times$ & $\times$ & $\times$ & 43.4& 28.1 \\
    $\checkmark$ & $\times$ & $\times$ & $\times$ & 45.3& 31.3 \\
    $\checkmark$ & $\checkmark$ & $\times$ & $\times$ & 47.0& 32.1 \\
    $\checkmark$ & $\checkmark$ & $\checkmark$ & $\times$ & 48.1 & 31.7\\ 
    $\checkmark$ & $\checkmark$ & $\checkmark$ & $\checkmark$ & 48.4& 32.3 \\
    \bottomrule 
  \end{tabular} 
  \end{adjustbox} 
  \caption{Effect of the proposed methods: CFM, DMA, PPC, and Refinement.}
  \label{tab:ablation_analysis}  
\end{table}
\begin{table}[t]
  \centering
  \begin{adjustbox}{max width=1\linewidth}
  \begin{tabular}{c|c|c|c}
    \toprule
    R & GFLOPs(G) & Memory(M) & {$mAP_{50}$} \\
    \midrule
    1 & 73.3 & 0.39 & 44.5 \\
    2 & 158.6 & 0.88 &  46.4 \\
    3 & 158.6 & 1.38 &  45.7 \\ 
    4 & 329.4 & 1.87 &  47.2 \\
    \bottomrule
  \end{tabular} 
  \vspace{-2mm}
  \label{table:RinCFM}
  \end{adjustbox}
  \caption{For the impact of different cascade numbers $R$ in CFM, we set the hidden layer dimension of CFM to 128.}
  \label{tab:R_analysis}
\end{table}
\begin{table}[t]
  \centering
  \begin{adjustbox}{max width=1\linewidth}
  \begin{tabular}{c|c|c|c}
    \toprule
    dim & GFLOPs(G) & Memory(M) & {$mAP_{50}$} \\
    \midrule
    32 & 23.9 & 0.13 & 44.8 \\
    64 & 86.7 & 0.48 & 46.9 \\
    128& 329.4 & 1.87 &  47.2 \\
    256 & 1282.5 & 7.37 &  46.7 \\ 
    \bottomrule
  \end{tabular} 
  \label{table:HidCFM}
  \end{adjustbox}
  \caption{For the impact of different hidden layer dimensions of CFM, we set the number of CFMs $R=4$.}
  \label{tab:dim_analysis}
\end{table}
\begin{table}[t]
  \centering
  \begin{adjustbox}{max width=1\linewidth}
  \begin{tabular}{c|c|c|c|c}
    \toprule
    backbone &dim& GFLOPs(G) & Memory(M) & {$mAP_{50}$} \\
    \midrule
    ResNet50${ }^{\ast}$ &32 & 55.9 & 26.1 &38.2 \\
    ResNet50${ }^{\ast}$ & 64& 62.3 & 26.8 & 48.4\\
    ResNet50${ }^{\ast}$ & 128 & 75.1 & 28.2 &  48.4 \\
    ResNet101${ }^{\ast}$ & 32 &68.8 & 45.1 &  39.1 \\ 
    HRNet-W34& 34 &63.7 & 30.5 &  44.3 \\ 
    HRNet-W48 & 48 &92.8 & 65.6 &  48.4 \\ 
    HRNetV2-W48 & 48 & 119.5& 69.8 & 48.0 \\ 
    \bottomrule
  \end{tabular} 
  \end{adjustbox}
   \caption{Analysis of the effect of Backbone result on our
WSIS performance.${ }^{\star}$ denotes the use of multi-scale backbone features.}
\label{tab:backbone_analysis}
\end{table}
\begin{table}[hbt!]
  \centering
  \label{tab:temperature_analysis}
  \setlength{\tabcolsep}{2pt} 
  \begin{adjustbox}{max width=1\linewidth}
  \begin{tabularx}{\linewidth}{@{}c|*{9}{X}@{}}
    \toprule
    Temperature & 0.1 & 0.2 & 0.3 & 0.4 & 0.5& 0.6 & 0.7 & 0.8 \\
    \midrule
    $mAP_{50}$ & 47.4 & 48.4 & 47.4 & 47.0 & 47.6 & 47.4& 47.7& 47.6\\
    \bottomrule
  \end{tabularx} 
 \end{adjustbox}
 \caption{Impact of Different Temperatures on Losses in the Pixel Contrast Module}
\label{tab:temperature_analysis}
\end{table}


The design of CFM helps decoupling the class-agnostic boundary detection branch from the semantic segmentation branch, while progressively extracting and refining boundary features across multiple stages. 
As shown in Table \ref{tab:ablation_analysis}, CFM leads to an improvement of 1.9\% in {$mAP_{50}$} and 3.2\% in {$mAP_{75}$}.
We also investigate the impact of the number and dimension of cascaded structures in CFM on GFLOPs, Memory, and {$mAP50$}. Table~\ref{tab:R_analysis} shows that the highest {$mAP_{50}$} is achieved when R equals 4, while Table~\ref{tab:dim_analysis} reveals that the best results are obtained when the dimension is set to 128. With these two hyper-parameters set, CFM strikes a balance between performance and computational efficiency.

\subsubsection{Impact of DMA}

The DMA is designed to capture weakly responsive instance boundary information from deeply fused features. As illustrated in the second row of Table~\ref{tab:ablation_analysis}, DMA leads to an improvement of 1.7\% in {$mAP_{50}$} and 0.8\% in {$mAP_{75}$}. As a submodule of CFM, DMA shares the optimal configuration with CFM.

\subsubsection{Impact of PPC}

The goal of Pixel-to-Pixel Contrast (PPC) is to enhance the perceptual capability of the IABD branch, with its core principle focusing on attracting similar pixels while repelling dissimilar ones. As indicated in the third row of Table~\ref{tab:ablation_analysis}, PPC leads to an improvement of 1.1\% in {$mAP_{50}$} and a decrease of 0.4\% in {$mAP_{75}$}. This decrease in {$mAP_{75}$} is attributed to the use of semantic contour pseudo-labels to sample pixels in PPC, resulting in rough predicted boundaries. We evaluate the impact of different temperatures on the weighted Pixel-to-Pixel Contrast loss. As shown in Table~\ref{tab:temperature_analysis}, the optimal performance is achieved when the temperature is set to $0.2$.

\subsubsection{Impact of Refinement}

The Mask Extraction Pipeline is designed to extract class-agnostic instance masks from instance boundaries. 
In particular, we observed that some instance masks generated by our pipeline exhibit issues such as holes and discontinuities in connectivity. Additionally, the incomplete closeness of instance boundaries leads to the diffusion of instance pixels into the background. The refinement procedure significantly ameliorates the above issues.
As shown in the last row of Table~\ref{tab:ablation_analysis}, the refinement results in a 0.3\% and 0.6\% improvement in terms of $mAP_{50}$ and $mAP_{75}$, respectively.

\subsubsection{Impact of Backbone}

The result in Table~\ref{tab:backbone_analysis} illustrates the impact of the Backbone selection on BAISeg performance. We employed HRNet-W48 \cite{tian2021learning} for BAISeg and achieved the best performance of 48.4 {$mAP_{50}$}.
An analysis of the first five rows of Table~\ref{tab:backbone_analysis} reveals that the performance of ResNet50 gradually improves with increasing dimension, reaching saturation at a dimension of 128. Given that the ResNet architectures require multi-scale fusion to match the performance of HRNet-W48, we consider HRNet-W48 more suitable for deployment.

\section{Conclusion}

This paper introduces BAISeg, a novel WSIS method that bridges the gap between semantic segmentation and instance segmentation via instance boundary detection. To decouple two parallel branches, we propose CFM. Within CFM, we introduced DMA to identify weak responsive instance boundaries. During training, PPC is employed to address issues like the non-closedness of instance boundaries. Our method demonstrates competitive results on both VOC 2012 and COCO datasets. Future work will focus on applying our approach to more challenging computer vision tasks, such as panoptic segmentation \cite{li2022mask}.

\newpage

\bibliographystyle{named}
\bibliography{ijcai24}

\clearpage
\appendix
\section{Problems with Popular WSIS Approaches}
Popular WSIS methods estimate a displacement field (DF) via learning inter-pixel relations and perform clustering to identify instances. However, the resulting instance centroids are inherently unstable~(even supervised by GT point labels) and vary significantly across different clustering algorithms.

\begin{figure}[ht!]
    \centering
    \includegraphics[width=\linewidth]{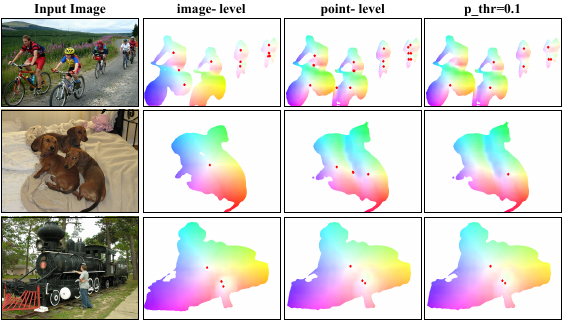}
    \caption{Illustration of the estimated Displacement Field and Centroids under different levels of annotation. ``p\_thr=0.1'' represents the Centroids generated after filtering with a threshold of 0.1 using pixel-level annotation. Visualization content is generated by BESTIE.}
    \label{fig:DF}
\end{figure}

To illustrate the problem more clearly, as shown in Figure \ref{fig:DF}, DF demonstrates instability when utilizing either image-level or point-level annotations. The instability of the centroid generated by clustering mainly reflects the following aspects. 
Firstly, a single centroid is often wrongly predicted for closely positioned instances of the same class. Sometimes the centroids of occluded instances are omitted. Secondly, sometimes multiple centroids are predicted for the same instance. Finally, it is hard to fix a single threshold to filter unreliable centroid.

To address the above limitations, we proposed Boundary-Assisted Instance Segmentation (BAISeg), which is a new paradigm that carries out instance segmentation by using pixel-level annotations only. The main idea is to pinpoint instance locations by leveraging instance boundaries, which can be effectively learned through the utilization of semantic-level boundaries. We designed the Instance-Assisted Boundary Detection (IABD) branch for this important task. Since instance boundary prediction is independent of centroid estimation, we effectively circumvent the issue of instability.

\section{Additional Ablation Study}

\subsection{Detailed study of components}
Here we provide some additional ablation studies. In Table ~3 of the full paper, we analyze the effectiveness of our approach and the performance improvements brought about by the combination of components.

As shown in Table~\ref{tab:ablation_analysis}, it can be found that compared with applying DMA and PPC alone, applying CFM can achieve better $mAP50$. It is worth noting that applying PPC alone does not bring any improvement. Likewise, combining DMA and PPC without CFM does not achieve the best performance. This is because of the coupling between the semantic segmentation branch and the boundary detection branch. It is ineffective to use the common features of the two branches to calculate the contrast between the pixels of the two categories. With the additional help of CFM, more independent edge features can be extracted for the IABD branch, and therefore better performance is achieved.
Experiments show that the combination of all of the three components reached the best performance.

\begin{table}[t]
  \centering
  \begin{adjustbox}{max width=1\linewidth}
  \input{ablation_analysis_more}
  \end{adjustbox} 
  \caption{Effect of the proposed methods: CFM, DMA, and PPC. Refinement is performed for all experiments.}
  \label{tab:ablation_analysis}  
\end{table}

\begin{table}[ht!]
  \begin{minipage}{0.48\linewidth} 
    \centering
    \begin{adjustbox}{max width=\linewidth}
      \input{ablation_ppc_pos}
    \end{adjustbox}
    \caption{Impact of embedded features from different cascade positions in CFM, as well as the effect of the weighted loss.}
    \label{tab:ppc_position}
  \end{minipage}\hfill
  \begin{minipage}{0.40\linewidth} 
    \centering
    \begin{adjustbox}{max width=\linewidth}
      \input{ablation_ppc_dim}
    \end{adjustbox}
    \caption{Impact of embedded features from different cascade positions in CFM on BAISeg.}
    \label{tab:ppc_embedding_dim}
  \end{minipage}
\end{table}

\begin{figure*}[ht!]
    \centering
    \includegraphics[width=\linewidth]{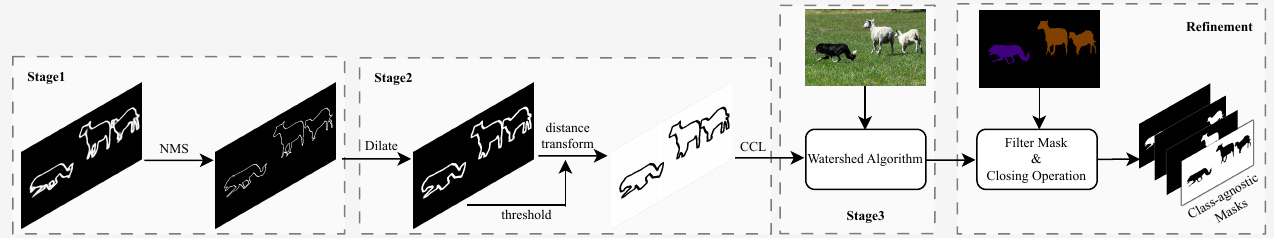}
    \caption{Mask Extraction Pipeline}
    \label{fig:MEP}
\end{figure*}

\subsection{Detailed study of PPC}
In the \textbf{section 3.4 Pixel-to-Pixel Contrast} of the full paper, we introduced PPC and Weighted Pixel-to-Pixel Contrast Loss.
First of all, we need to understand that the embedded features $\mathcal{F}_p$ used by PPC come from CFM. As CFM is composed of R cascade structures, the source of embedded features can be extracted from any one of the cascade structures. Therefore, we explore the impact of the different cascade structure positions where we extract embedding features from on the performance. As shown in Table \ref{tab:ppc_position}, we conducted experiments on embedding features from different cascaded structures and found that when the features are extracted from the second cascaded module (i.e. $R=2$), the performance of BAISeg reaches the best. 

Besides, we also explored the impact of the projected feature dimensions on performance. As shown in Table~\ref{tab:ppc_embedding_dim}, the best performance is achieved when the projected feature dimension is $1$. 

Finally, the impact of Weighted Pixel-to-Pixel Contrast Loss on performance is explored. As shown in  Table~\ref{tab:ppc_position}, under the same configuration, when using the weighted loss (the third row of the table), $mAP50$ reaches $48.4$, otherwise dropping to $47.0$. This indicates that Weighted Pixel-to-Pixel Contrast Loss can effectively mitigate the problem of semantic drift.

\begin{figure}[t]
    \centering
    \includegraphics[width=\linewidth]{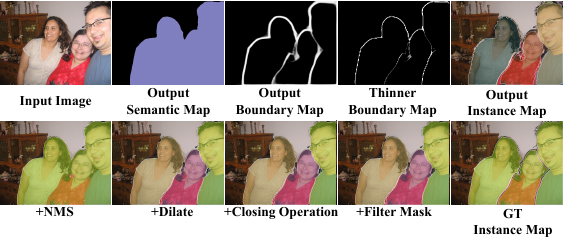}
    \caption{Illustration of the impact of adding different morphological components on instance segmentation prediction. The data comes from the validation set of PASCAL VOC 2012.}
    \label{fig:visual_MEP}
\end{figure}

\section{Detail of Mask Extraction Pipeline}
BAISeg relies on the IABD branch to output instance-aware boundary maps and further uses the Mask Extraction Pipeline (MEP) to obtain class-independent instance masks. In simple terms, we utilize the well-known watershed algorithm to extract instance masks from the boundary map. However, the instance boundaries generated by the IABD branch have the following issues: (1) roughness of the boundaries, (2) closedness of the boundaries, and (3) holes in the masks. Therefore, we propose MEP to address these problems. Here we introduce the detailed structure of MEP.

As shown in Figure~\ref{fig:MEP}, the process of the pipeline is detailed. The whole pipeline can be divided into three stages. In stage one, Non-Maximum Suppression (NMS) is applied to obtain a refined thinner boundary map. Specifically, NMS is used to suppress redundant edges to more accurately delineate the positions of boundary pixels.

In stage two, the goal is to extract a label map from the instance boundaries using the connected component labeling (CCL) algorithm, which includes dilated operation, distance transform operation, and threshold operation. Specifically, to further ensure the smoothness and closure of instance boundaries, a dilated operation with \text{iterations=2} is applied to the thin boundary map. Subsequently, the dilated boundary map is processed by the distance transform algorithm and threshold processing to generate a denoised distance map, and the label map is obtained via CCL.

In the third stage, the input image and label map are pushed to the watershed algorithm to obtain a collection of class-agnostic masks. We further observed that the class-agnostic instance masks extracted through the above steps have flaws. For instance,  holes may present inside instance masks, and pixels in some instances might significantly diffuse beyond their original semantic area. 

To fill the holes in the derived instance masks, we further employed a closing operation.

The closing operation employs a $(5,5)$ elliptical structure to perform dilation followed by erosion on the masks. This process aids in enhancing the connectivity of the masks, filling in small holes, and eliminating minor noise masks.
To improve the closedness of the boundaries, the semantic map is utilized as a filter to select only the pixels within the semantic region. We refer to the above additional operations as ``refinement''.

We have provided additional visualizations for the morphological methods in MEP, as shown in Figure \ref{fig:visual_MEP}. In the first row, we visualize the outputs of BAISeg and the instance segmentation results obtained without incorporating any morphological methods. In the second row, we gradually add morphological processes under the initial MEP setup, which significantly enhances the quality of the instance segmentation masks.

\begin{table*}[t]
	\centering
	\renewcommand\arraystretch{1.2}
	\setlength{\tabcolsep}{9pt}
	\vspace{-1mm}
	\resizebox{\linewidth}{!}{%
		\input{arch_exp_single}
            }
 \caption{\textbf{Single-branch approach with only the IABD branch.} Semantic and instance segmentation performance on the validation set of PASCAL VOC 2012. The three left columns correspond to the semantic segmentation methods we explored to provide semantic maps for training the IABD branch. ``Superv'' represents the training supervision ($\mathcal{P_M}$: Pixel-level Mask, $\mathcal{I}$: Image-level class label). The backbone we used for BAISeg is HRNet-W48. 
 ``Inference\_SM'' shows the semantic segmentation methods that provide the semantic maps. ``RCNN'' denotes additional training with Mask R-CNN to refine the prediction. ``Semantic\_gd'' means the ground truth of semantic segmentation. 
 }
 \label{tab:arch_single}
\end{table*}

\begin{table*}[ht!]
	\centering
	\renewcommand\arraystretch{1.2}
	\setlength{\tabcolsep}{9pt}
	\vspace{-1mm}
	\resizebox{\linewidth}{!}{%
		\input{arch_exp_double}
            }
 \caption{\textbf{Double-branch approach with IABD branch and semantic segmentation branch.} Semantic and instance segmentation performance on the validation set of PASCAL VOC 2012. The three left columns correspond to the semantic segmentation methods we explored to provide semantic maps for training the IABD branch and semantic segmentation branch. ``Superv'' represents the training supervision ($\mathcal{P_M}$: Pixel-level Mask, $\mathcal{I}$: Image-level class label). The backbone we used for BAISeg is HRNet-W48.
 ``Inference\_SM'' shows the semantic segmentation methods that provide the semantic maps. ``RCNN'' denotes additional training with Mask R-CNN to refine the prediction. ``Semantic\_gd'' means the ground truth of semantic segmentation. BAISeg is our Semantic Segmentation Branch Prediction.
 }
 \label{tab:arch_double}
\end{table*}

\begin{figure}[ht!]
    \centering
    \includegraphics[width=\linewidth]{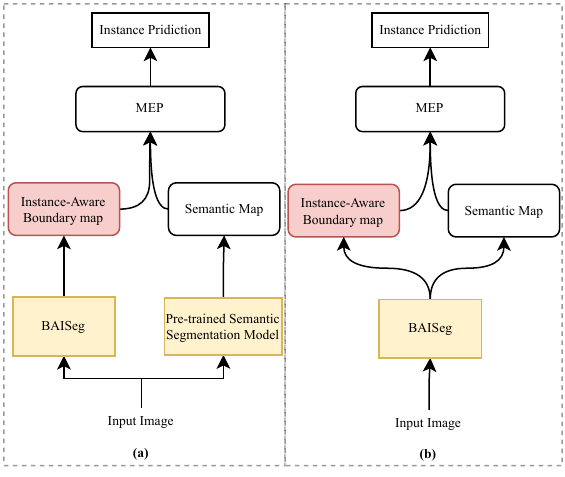}
    \caption{BAISeg Architectures.}
    \label{fig:arch}
\end{figure}
\section{Architecture exploration}


BAISeg is a very flexible method. As shown in Figure~\ref{fig:arch}, BAISeg can be applied in two ways in practice, which we call the single-branch approach and the double-branch approach.

The single-branch approach is shown in detail in Figure~\ref{fig:arch} (a). The input image is transmitted to a pre-trained semantic segmentation model and BAISeg respectively.
The semantic map is obtained by the semantic segmentation model. BAISeg uses the IABD branch to estimate the boundaries of instances. Finally, instance boundaries and semantic segmentations are combined to derive instance segmentations.

Figure~\ref{fig:arch} (b) shows the double-branch approach in detail. The input image is fed to BAISeg. The two parallel branches of BAISeg simultaneously generate semantic maps and instance-aware boundary maps and are processed by the MEP for instance segmentation.

\subsection{The Single-Branch Experiments}

Different from the dual-branch training shown in Figure~2 of the full paper, in the single-branch method, BAISeg only predicts the instance-aware boundary maps by training the IABD branch to further discriminate the instances.

Notably, the semantic contour labels trained by the IABD branch are extracted from segmentation masks using gradient derivation. To explore the potential of segmentation masks of different qualities and methods relative to single-branch methods, we utilize segmentation masks generated by popular pre-trained segmentation models for training. As shown in the first column of Table~\ref{tab:arch_single}, pre-trained segmentation models include PPM, FCN, Deeplabv3, and OCRNet.

We explore the performance changes of segmentation masks of different qualities used for training the IABD branch in the single-branch approach. As shown in Table \ref{tab:arch_single}, as the quality of segmentation masks increases, $mAP50$ also increases, because the quality of segmentation masks directly affects the accuracy of instance pseudo-boundaries. It is worth noting that although PPM has a higher mIoU value, it achieves poorer $mAP50$ performance. This is because PPM uses image-level annotations, so it performs worse in perceiving mask boundaries. 

We also studied the performance when using segmentation masks from DeepLabv3 for IABD training. When combined with various segmentation masks, we found that the performance using Semantic\_gd for inference is the best. As shown in the last two rows of Table~\ref{tab:arch_single}, more experiments are done on Semantic\_gd to explore the potential of single-branch training.

\subsection{The Double-Branch Experiments}

Figure~2 in the full paper shows the architecture of dual-branch training in detail. Here, the same settings as for the training of the single-branch approach are kept.
Both BAISeg's branches are trained using segmentation masks either generated by the selected pre-trained segmentation models or ground truth. Pretrained models include PPM, FCN, Deeplabv3, and OCRNet.

We explore the performance changes of segmentation masks of different qualities used for training the BAISeg in the double-branch approach. As shown in Table~\ref{tab:arch_double}, as the quality of segmentation masks increases, the performance also increases. As in the single-branch approach, the approach using image-level labels still obtains poor performance despite the higher IoU of the segmentation mask.

Similar to the case of the single-branch approach, we also tested the performance when using segmentation masks from OCRNet for IABD training. When combined with various segmentation masks for inference, we found that the performance using Semantic\_gd is the best.

The last three rows of the table show more experimental results on Semantic\_gd to explore the potential of dual-branch training.

\section{Definition of ResNet multi-scale features}
In the subsection \textbf{Impact of Backbone} of section 4.4, we explored the impact of different backbone networks on BAISeg. We used multi-scale ResNet as the backbone network for experiments, whose detail is shown as follows. 

In particular, we start by using ResNet-50/101 to extract a set of feature maps. The output of the ResNet is then used to construct the feature pyramid $\left\{P_1, P_2, P_3, P_4, P_5\right\}$. We apply a $1 \times 1$ convolution to each feature map individually, uniformly mapping to the same dimension, reducing it to a pre-defined smaller dimension if the channel dimension is larger than that. Then these features are concatenated together to formulate the final feature map. We formulate this procedure as:

\begin{equation}
P_{\text {cat }}=\operatorname{Cat}\left[\mathrm{E}\left(P_1\right);\mathrm{E}\left(P_2\right);\mathrm{E}\left(P_3\right)\uparrow ; \mathrm{E}\left(P_4\right) \uparrow ; \mathrm{E}\left(P_5\right) \uparrow\right],
\end{equation}
where $\mathrm{E}(\cdot)$ represents feature embedding operation $(1 \times 1$ convolution). Cat $[\cdot ; \cdot]$ means channelwise concatenation. $\uparrow$ denotes the upsampling operation. 




\section{Qualitative Results of BAISeg}

In Figure~\ref{fig:coco_qulization}, we provide more qualitative results of BAISeg with the double-branch approach.
The output of BAISeg includes semantic segmentation mask and instance boundary detection. MEP is then used to extract class-independent masks from the boundary maps of instances, and outputs for instance segmentation are obtained by combining the semantic maps and instance masks. To obtain refined and high-quality masks, we use Mask-RCNN for Refinement.
Furthermore, we compare our instance masks with those of CIM, which is a proposal-based approach. The comparison results clearly show that our approach can properly segment multiple instances with a high-precision instance mask.

\begin{figure}[t]
    \centering
    \includegraphics[width=\linewidth]{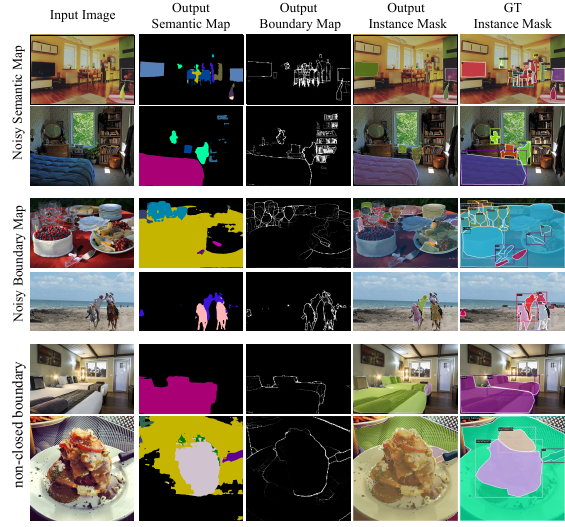}
    \caption{ Failure cases of BAISeg.}
    \label{fig:failure_cases}
\end{figure}
\begin{figure*}[t]
    \centering
    \includegraphics[width=\linewidth]{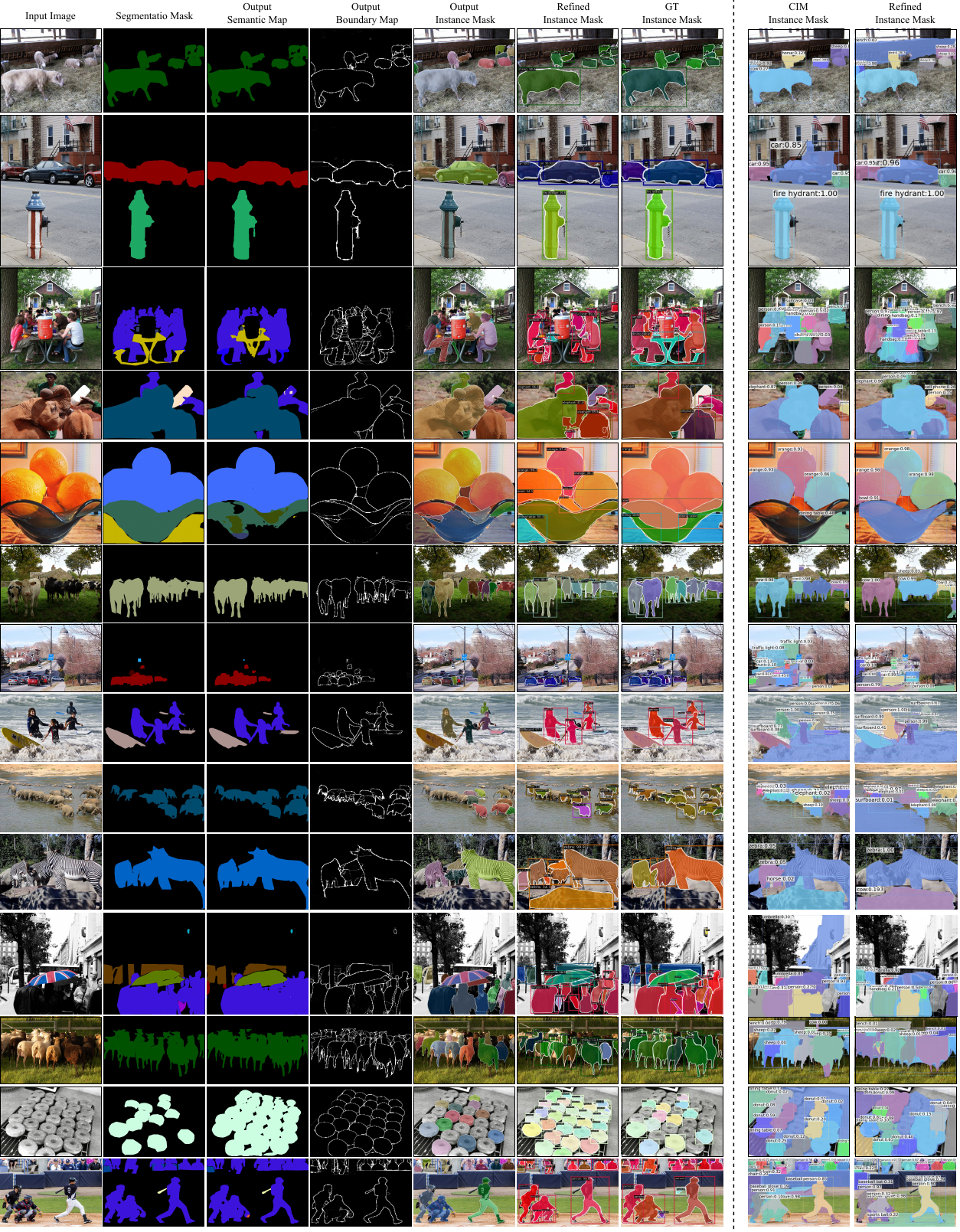}
    \caption{Qualitative results of our BAISeg and outputs of CIM on COCO 2017 dataset. ``Segmentation Mask'' corresponds to the labels used for training. ``Output Semantic Map'' shows the segmentation masks predicted by the semantic segmentation branch. ``Output Boundary Map'' shows the instance boundaries predicted by the IABD branch. ``Output Instance Mask'' illustrates the instance masks generated by combining the results of both branches, whereas ``Refined Instance Mask'' shows the instance masks after refinement. The boundary map here is processed by NMS. Note that we only use pixel-level labels without off-the-shelf proposal techniques. Compared with CIM which is the proposal method, our BAISeg can segment multiple instances more accurately and precisely.}
    \label{fig:coco_qulization}
\end{figure*}

\section{Limitations}

Although BAISeg achieves good results in performance, there is still more room for improvement. 
First, the instance boundaries estimated under semantic contour labels tend to be coarser, further affecting the performance of BAISeg.
Second, the instance boundaries predicted by BAISeg are sometimes not closed, which will cause multiple instances to be merged into single instances.

We provide some failure cases of BAISeg in Figure~\ref{fig:failure_cases}.
First, when the semantic segmentation map provides a noisy foreground region, we often fail to obtain the precise instance mask (first row in Figure~\ref{fig:failure_cases}). 
In addition, noisy Boundary maps lead to false instance masks (second row in Figure~\ref{fig:failure_cases}). Last, when the Instance-Aware Boundary Detection branch provides an unclosed boundary map, we often fail to obtain the precise instance mask (last row in Figure~\ref{fig:failure_cases}).

\end{document}

%% file: ablation_analysis_more.tex
\begin{tabular}{ccc|S[table-format=2.1]S[table-format=2.1]}
    \toprule
    CFM & DMA & PPC & {$mAP_{50}$} & {$mAP_{75}$}  \\
    \midrule
    $\checkmark$ & $\times$ & $\times$& 45.8& 31.9 \\
    $\times$ & $\checkmark$ & $\times$& 45.1& 31.1 \\
    $\times$ & $\times$ & $\checkmark$&43.4 & 30.1\\ 
    $\checkmark$ & $\checkmark$ & $\times$&  47.2& 32.9 \\
    $\checkmark$ & $\times$ & $\checkmark$ & 46.2& 32.4 \\
    $\times$ & $\checkmark$ & $\checkmark$ &45.3& 31.8 \\
    $\checkmark$ & $\checkmark$ & $\checkmark$ & 48.4 & 32.3 \\
    \bottomrule 
  \end{tabular} 

%% file: ablation_ppc_pos.tex
\begin{tabular}{c|c|S[table-format=2.1]}
    \toprule
    R & weighted Loss & {$mAP_{50}$}  \\
    \midrule
    1 & $\checkmark$ & 47.4 \\
    2 & $\times$  & 47.0 \\ 
    2 & $\checkmark$  & 48.4 \\ 
    3 & $\checkmark$ & 46.9 \\
    4 & $\checkmark$  & 47.5 \\
    \bottomrule 
  \end{tabular} 

%% file: ablation_ppc_dim.tex
\begin{tabular}{c|S[table-format=2.1]}
    \toprule
    dimention &{$mAP_{50}$}  \\
    \midrule
    1 &  48.4 \\
    32 &  47.3 \\
    64 &  46.9 \\ 
    128 &  47.1 \\ 
    \bottomrule 
  \end{tabular} 

%% file: arch_exp_single.tex
\begin{tabular}{c | c | c | c | c c  c  c  c  c}
\hline
Train\_{SM}  & Superv &mIoU& Inference\_{SM}&Superv& RCNN &{$mAP_{25}$} & {$mAP_{50}$}& {$mAP_{70}$} & {$mAP_{75}$}\\
\hline
PPM           &$\mathcal{I}$&70.0&PPM& $\mathcal{P_M}$& $\times$ & 43.5 & 35.8  & 25.2  & 22.0 \\    
FCN            &$\mathcal{P_M}$&66.9&FCN & $\mathcal{P_M}$& $\times$ & 46.4  & 38.6 & 28.7 & 25.4\\
OCRNet  &$\mathcal{P_M}$&77.1&OCRNet  & $\mathcal{P_M}$& $\times$ & 50.3  & 42.1 & 31.8 & 28.5\\
\hline
DeepLabv3 &$\mathcal{P_M}$&77.6&DeepLabv3  & $\mathcal{P_M}$& $\times$ & 51.5  & 42.5 & 32.4 & 28.9\\
DeepLabv3 &$\mathcal{P_M}$&77.6&PPM  & $\mathcal{P_M}$& $\times$ & 46.3  & 39.1 & 30.5 & 27.3\\
DeepLabv3 &$\mathcal{P_M}$&77.6&FCN  & $\mathcal{P_M}$& $\times$ & 49.4  & 40.4 & 30.0 & 26.5\\
DeepLabv3 &$\mathcal{P_M}$&77.6&OCRNet  & $\mathcal{P_M}$& $\times$ & 51.0  & 42.3 & 32.5 & 28.9\\
DeepLabv3 &$\mathcal{P_M}$&77.6&Semantic\_gd  & $\mathcal{P_M}$& $\times$ & 55.6  & 47.7 & 37.1 & 33.4\\
\hline 
Semantic\_gd                       &--&--&Semantic\_gd& $\mathcal{P_M}$& $\times$ & 56.9  & 50.6 & 40.2 & 36.4\\
Semantic\_gd                       &--&--&Semantic\_gd& $\mathcal{P_M}$& $\checkmark$ & \textbf{66.1}  & \textbf{61.5} & \textbf{45.1} & \textbf{36.0}\\
\hline

\end{tabular}

%% file: arch_exp_double.tex
\begin{tabular}{c | c | c | c | c c  c  c  c  c}
\hline
Train\_{SM}  & Superv &mIoU& Inference\_{SM}&Superv & RCNN &{$mAP_{25}$} & {$mAP_{50}$}& {$mAP_{70}$} & {$mAP_{75}$}\\
\hline

PPM           &$\mathcal{I}$&70.0&BAISeg& $\mathcal{P_M}$& $\times$ & 44.3  & 35.8 & 25.5 & 22.0 \\
FCN            &$\mathcal{P_M}$&66.9&BAISeg & $\mathcal{P_M}$& $\times$ & 49.6  & 40.3 & 30.1 & 26.4\\
DeepLabv3 &$\mathcal{P_M}$&77.6&BAISeg  & $\mathcal{P_M}$& $\times$ & 53.0  & 44.0 & 33.0 & 29.5\\
\hline
OCRNet  &$\mathcal{P_M}$&77.1&BAISeg  & $\mathcal{P_M}$& $\times$ & 52.7  & 44.0 & 32.1 & 28.9\\
OCRNet  &$\mathcal{P_M}$&77.1&OCRNet  & $\mathcal{P_M}$& $\times$ & 52.4  & 43.7 & 32.8 & 28.9\\
OCRNet  &$\mathcal{P_M}$&77.1&PPM  & $\mathcal{P_M}$& $\times$ & 47.8  & 39.9 & 30.4 & 26.9\\ 
OCRNet  &$\mathcal{P_M}$&77.1&FCN  & $\mathcal{P_M}$& $\times$ & 51.1  & 41.9 & 30.5 & 27.2\\
OCRNet  &$\mathcal{P_M}$&77.1&DeepLabv3  & $\mathcal{P_M}$& $\times$ & 52.9  &44.2 & 33.1 & 29.3\\
OCRNet  &$\mathcal{P_M}$&77.1&Semantic\_gd & $\mathcal{P_M}$ & $\times$ & 56.2  & 47.7 & 36.8 & 33.5\\
\hline 
Semantic\_gd                       &--&--&BAISeg& $\mathcal{P_M}$& $\times$ & 55.3  & 48.4 & 36.9 & 32.3\\
Semantic\_gd                       &--&--&Semantic\_gd& $\mathcal{P_M}$& $\times$ & 59.2  & 53.0 & 42.0 & 37.6\\
Semantic\_gd                       &--&--&Semantic\_gd& $\mathcal{P_M}$& $\checkmark$ & \textbf{69.4}  & \textbf{62.0} & \textbf{44.0} & \textbf{36.0}\\
\hline

\end{tabular}